\documentclass[letterpaper, 10 pt, journal, twoside]{IEEEtran}

\usepackage{amsmath,amsfonts}
\usepackage{algorithmic}
\usepackage{algorithm}
\usepackage{array}
\usepackage{textcomp}
\usepackage{stfloats}
\usepackage{url}
\usepackage{verbatim}
\usepackage{graphicx}
\usepackage{cite}

\usepackage[dvipsnames]{xcolor}
\usepackage{xcolor}
\usepackage{soul}
\usepackage[printonlyused]{acronym}
\usepackage{hyperref}
\hypersetup{
    colorlinks=true,
    linkcolor=red,
    filecolor=magenta,      
    urlcolor=cyan,
    pdftitle={Overleaf Example},
    pdfpagemode=FullScreen,
    }

\usepackage{subfigure}
\usepackage{comment}
\usepackage{multirow}
\usepackage{float}
\usepackage{siunitx}
\usepackage{amssymb}
\usepackage{bm}
\usepackage{nicefrac}
\usepackage[placement=top,vshift=-7]{background}

\usepackage{tikz}
\usepackage{pgfplots}
\pgfplotsset{compat=1.14}
\usetikzlibrary{backgrounds,arrows,automata,shapes,positioning,calc,through,spy}
\pgfdeclarelayer{background}
\pgfdeclarelayer{foreground}
\pgfsetlayers{background,main,foreground}

\tikzset{
  state/.style={
    rectangle,
    draw=black, very thick,
    minimum height=1.0em,
    text centered,
  },
  smallstate/.style={
    rectangle,
    draw=black, very thick,
    minimum height=0.2em,
    text centered,
  },
  final_state/.style={
    rectangle,
    rounded corners,
    draw=black, very thick,
    minimum height=2em,
    text centered,
  },
  initial_state/.style={
    rectangle,
    double=white,
    double distance=1pt,
    inner sep=2pt,
    draw=black, very thick,
    minimum height=2em,
    text centered,
  },
  color_state/.style={
    rectangle,
    draw=black, very thick,
    fill=yellow!40,
    minimum height=2em,
    text centered,
  },
  point/.style={
    circle,
    inner sep=0pt,
    minimum size=3pt,
    fill=red
  },
  adder/.style={
    circle,
    inner sep=2pt,
    minimum size=0.3in,
    draw=black, very thick,
    text centered
  },
  state_gray/.style={
    rectangle,
    draw=black, very thick,
    fill=gray!40,
    minimum height=2.0em,
    text centered,
  },
  state_white/.style={
    rectangle,
    draw=black, very thick,
    fill=white,
    minimum height=1.0em,
    text centered,
    text=black,
    inner sep=0,
  },
  state_green/.style={
    rectangle,
    draw=black, very thick,
    fill=green!50,
    minimum height=1.0em,
    text centered,
    text=black,
    inner sep=0,
  },
  state_red/.style={
    rectangle,
    draw=black, very thick,
    fill=red!70,
    minimum height=1.0em,
    text centered,
    text=black,
    inner sep=0,
  },
  state_org/.style={
    rectangle,
    draw=black, very thick,
    fill=orange!40,
    minimum height=1.0em,
    text centered,
    text=black,
    inner sep=0,
  },
  state_blue/.style={
    rectangle,
    draw=black, very thick,
    fill=blue!40,
    minimum height=1.0em,
    text centered,
    text=black,
    inner sep=0,
  },
  final_state/.style={
    rectangle,
    rounded corners,
    draw=black, very thick,
    minimum height=2em,
    text centered,
  },
  initial_state/.style={
    rectangle,
    double=white,
    double distance=1pt,
    inner sep=2pt,
    draw=black, very thick,
    minimum height=2em,
    text centered,
  },
  point/.style={
    circle,
    inner sep=0pt,
    minimum size=3pt,
    fill=red
  },
}

\tikzset{new spy style/.style={spy scope={%
  magnification=5,
  size=1.25cm,
  connect spies,
  every spy on node/.style={
    rectangle,
    draw,
  },
  every spy in node/.style={
    draw,
    rectangle,
    fill=white
  }
  }
  }
}

\begin{document}
\SetBgScale{1.0}
\SetBgContents{IEEE ROBOTICS AND AUTOMATION LETTERS. PREPRINT VERSION - DO NOT DISTRIBUTE. \href{https://doi.org/10.1109/LRA.2022.3231831}{DOI 10.1109/LRA.2022.3231831}}
\SetBgColor{black}
\SetBgAngle{0}
\SetBgOpacity{1.0}
\newcommand{\usvaxis}{b}
\newcommand{\discusvaxis}{b}
\newcommand{\usvstatevect}{\mathbf{v}}
\newcommand{\usvstatevectder}{\dot{\usvstatevect}}
\newcommand{\discusvstatevect}{\mathbf{v}}
\newcommand{\discusvstatevectder}{\dot{\mathbf{v}}}
\newcommand{\phaseji}{\phi_{j,i}}
\newcommand{\Phaseji}{\Phi_{j,i}(t) }
\newcommand{\phase}{\phi}
\newcommand{\Phase}{\Phi}
\newcommand{\obsPhase}{\Phase_{j,i}(t_{obs})}
\newcommand{\ampji}{A_{j,i}}
\newcommand{\amp}{A}
\newcommand{\obsAmp}{\amp_{j,i}(t_{obs})}
\newcommand{\freqji}{f_{j,i}}
\newcommand{\freq}{f}
\newcommand{\fftdelta}{\Delta T_{FFT}}
\newcommand{\discsamplingtime}{\Delta T}
\newcommand{\identime}{t_{FFT}}
\newcommand{\predictedstatevect}{\hat{\statevect}}
\newcommand{\outputvect}{\mathbf{y}}
\newcommand{\singlestatematrix}{\mathbf{B}}
\newcommand{\discstatematrix}{\mathbf{\Psi}}
\newcommand{\singleoutputmatrix}{\mathbf{C}}
\newcommand{\combinedstatematrix}{\overline{\singlestatematrix}}
\newcommand{\combinedoutputmatrix}{\overline{\singleoutputmatrix}}
\newcommand{\kalmangain}{\mathbf{L}}
\newcommand{\systemnoisemat}{\mathbf{Q}}
\newcommand{\observationnoise}{\mathbf{R}}
\newcommand{\systemnoisematI}{\mathbf{Q_I}}
\newcommand{\processcovariancemat}{\mathbf{P}}
\newcommand{\dronestatevector}{\mathbf{x}}
\newcommand{\dronestatevectdes}{\overset{*}{\dronestatevector}}
\newcommand{\errinz}{\Tilde{z}}
\newcommand{\droneinputvector}{\mathbf{u}}
\newcommand{\dronestatematrix}{\mathbf{D}}
\newcommand{\droneinputmatrix}{\mathbf{E}}
\newcommand{\deltapred}{\Delta t_{pred}}
\newcommand{\substatematrix}{\mathbf{D'}}
\newcommand{\subinputmatrix}{\mathbf{E'}}
\newcommand{\heading}{\eta}
\newcommand{\objfunction}{J(\dronestatevector, \droneinputvector)}
\newcommand{\errormat}{\mathbf{\Tilde{\dronestatevector}}}
\newcommand{\errpenmat}{\mathbf{S}}
\newcommand{\inputeffortvect}{\mathbf{h}}
\newcommand{\inputeffortpen}{\mathbf{T}}
\newcommand{\predhorizon}{M_p}
\newcommand{\controlhorizon}{M_c}
\newcommand{\medialink}{http://mrs.felk.cvut.cz/ral-landing-on-usv}

\acrodef{uav}[UAV]{unmanned aerial vehicle}
\acrodef{mpc}[MPC]{model predictive control}
\acrodef{nmpc}[NMPC]{nonlinear model predictive control}
\acrodef{usv}[USV]{unmanned surface vehicle}
\acrodef{fft}[FFT]{fast Fourier transform}
\acrodef{ode}[ODE]{ordinary differential equation}
\acrodef{dof}[DOF]{degrees of freedom}
\acrodef{imu}[IMU]{inertial measurement unit}

\def\changehglt{1}
\newcommand{\RT}{ForestGreen}
\newcommand{\RE}{blue}
\newcommand{\RTT}{orange}

\title{Landing a UAV in harsh winds and turbulent open waters}

\author{Parakh~M.~Gupta, Èric Pairet, \IEEEmembership{Member,~IEEE,} Tiago Nascimento, \IEEEmembership{Senior Member,~IEEE,} and Martin Saska, \IEEEmembership{Member,~IEEE}
\thanks{Manuscript received: August 16, 2022; Revised: November 11, 2022; Accepted: December 5, 2022. This paper was recommended for publication by Editor Pauline Pounds upon evaluation of the Associate Editor and Reviewers’comments. \textit{(Corresponding author: Parakh~M.~Gupta)}}
\thanks{This work has been supported by the Technology Innovation Institute - Sole Proprietorship LLC, UAE, under the Research Project Contract No. TII/ARRC/2055/2021, by CTU grant no SGS20/174/OHK3/3T/13, by the Czech Science Foundation (GAČR) under research project No. 20-10280S, by the European Union’s Horizon 2020 research and innovation program AERIAL-CORE under grant agreement no. 871479, and by the Ministry of Education of the Czech Republic through the OP VVV funded project CZ.02.1.01/0.0/0.0/16 019/0000765 ``Research Center for Informatics".}
\thanks{P. M. Gupta, T. Nascimento, and M. Saska are with the Department of Cybernetics, Czech Technical University in Prague, Prague, Czech Republic (e-mail: guptapar@fel.cvut.cz, see http://mrs.felk.cvut.cz)}
\thanks{T. Nascimento is also with the Department of Computer Systems, Universidade Federal da Paraíba, Brazil}
\thanks{È. Pairet is with the Technology Innovation Institute, Abu Dhabi (UAE)}
\thanks{Digital Object Identifier (DOI): see top of this page.}
}

\markboth{IEEE ROBOTICS AND AUTOMATION LETTERS. PREPRINT VERSION. ACCEPTED DECEMBER, 2022 }
{Gupta \MakeLowercase{\textit{et al.}}: Landing a UAV in harsh winds and turbulent open-waters}

\graphicspath{{./figures/}}

\maketitle

\begin{abstract}
Landing an \ac{uav} on top of an \ac{usv} in harsh open waters is a challenging problem, owing to forces that can damage the \ac{uav} due to a severe roll and/or pitch angle of the \ac{usv} during touchdown. To tackle this, we propose a novel \ac{mpc} approach enabling a \ac{uav} to land autonomously on a \ac{usv} in these harsh conditions. The \ac{mpc} employs a novel objective function and an online decomposition of the oscillatory motion of the vessel to predict, attempt, and accomplish the landing during near-zero tilt of the landing platform. The nonlinear prediction of the motion of the vessel is performed using visual data from an onboard camera. Therefore, the system does not require any communication with the \ac{usv} or a control station. The proposed method was analyzed in numerous robotics simulations in harsh and extreme conditions and further validated in various real-world scenarios.
\end{abstract}


\section{Introduction}
\IEEEPARstart{H}{eterogeneous} robot teams that are composed of \acp{uav} and \acp{usv} are aimed to provide higher efficiency and decrease the high risk posed to human life in marine applications. An example of such an application is the process of cleaning oceans to rid them of oil spills and non-biodegradable waste \cite{lebreton2017river}. While the \acp{uav} can act as the eyes in the sky for surveying, identifying, and localizing the clean-up targets, the \acp{usv} are much better suited to the actual clean-up as this task requires heavy equipment and lifting capabilities close to the water surface. 

These clean-up missions can be performed autonomously by \acp{uav} and can be conducted several dozen kilometers away from a harbor or shore. Although \acp{uav} have short battery lives to be able to fly long distances, their strength lies in their agility and their ability to perform short-duration hover missions \cite{nascimento2019position}. We can compensate for this short battery life by making a \ac{uav} and \ac{usv} behave as a team, where-in the \ac{uav} can charge quickly during the mission for rapid redeployment. However, the precipitous and violent nature of the sea poses daunting challenges for landing on the \ac{usv} deck, especially due to the precision required for recharging operations. 

\begin{figure}[!t]
    \centering
    \includegraphics[width=0.3\textwidth]{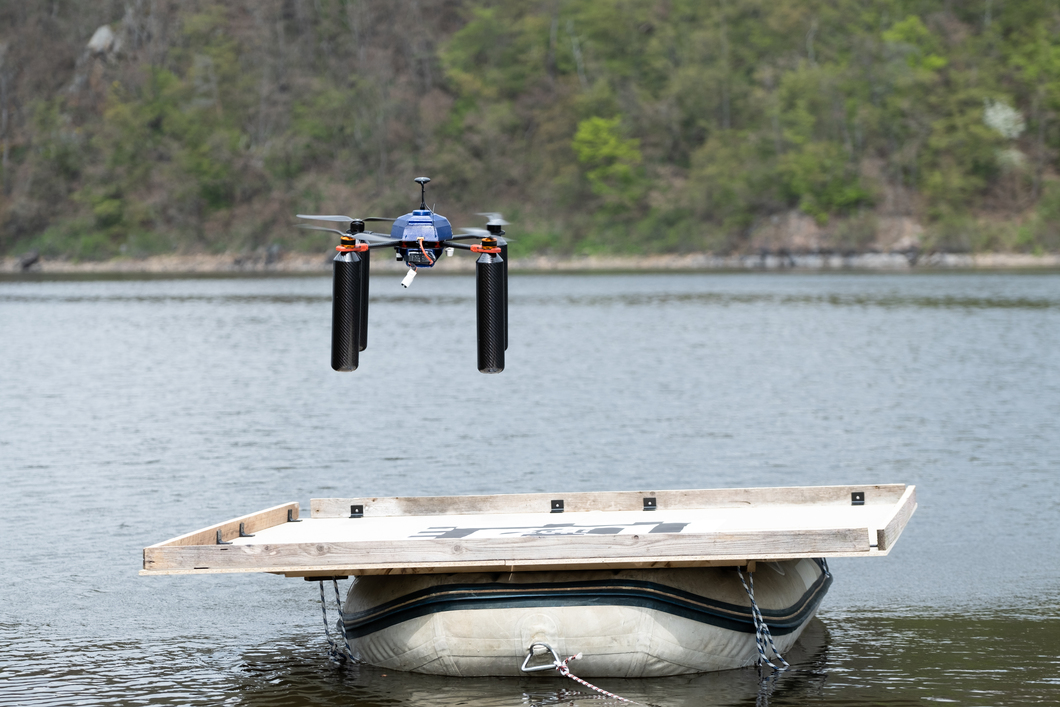}
    \caption{\ac{uav} landing on \ac{usv} in real-world experiments.}
    \label{fig:landing_image_real}
\end{figure}

When landing on a \ac{usv}, the first challenge is estimating and predicting the movement of the deck of the \ac{usv} before landing. A fast-moving deck can damage the \ac{uav} during landing through high impulse impacts, while a tilted deck can result in the \ac{uav} rolling or falling off the deck before the landing is complete. Additionally, a tilted deck can cause an erroneous response from the controller of the \ac{uav} during landing, which would jeopardize the landing position since multi-rotors are under-actuated vehicles with coupled angular and linear acceleration vectors. The second challenge that we focus on is attempting a landing without active communication between the \ac{uav} and the \ac{usv}. Relying on a required communication channel with a high frame rate and low latency would introduce a significant source of failure in real open-water applications.To increase reliability and applicability, we attempt to build a decentralized solution that does not rely on communication between the agents. Thus, we aim to study various aspects of the dynamics of \acp{uav} and \acp{usv} to develop a framework for predicting and landing on the \ac{usv} with high precision and reliability in demanding conditions including wind and waves, often seen in harsh environments. Finally, in this work, we can define harsh environments as those that contain open water with waves with a height of up to 4 meters, and a wind velocity of up to 12m/s, which corresponds to a Beaufort scale of 6. For intended applications, this would produce a tilt in the range of [-0.5,0.5] radians for the \ac{usv}.

\section{Related Works}
\label{related_works}

Riola et al. \cite{Riola2011} show that the behavior of a ship can be predicted based on its past motion up to short prediction horizons if corrected by measured ship motion. Unsurprisingly, the topic of wave predictions is highly relevant to the shipping industry as it is needed to prevent cables from slacking while trying to offload cargo from ships using port-side cranes. Both K\"{u}chler et al. \cite{Kuchler2011} and Neupert et al. \cite{Neupert2008} describe an active heave compensation for port-side cranes using a periodic oscillation model that proves to be effective. 

Building on a similar model, Marconi et al. \cite{Marconi2002} and Lee et al. \cite{Lee2019} present sophisticated approaches for fixed-wing \acp{uav} landing on vessels. Both works adapt the model using a Kalman filter and use this heave motion of the ship to predict the altitude of the landing pad. However, these works do not focus on a rolling and pitching deck. Meng et al. \cite{Meng2019} take a different approach and use an auto-regressive-model on the fixed-wing \ac{uav} to observe and predict the ship motion by breaking it into sinusoidal components. In addition, Ngo and Sultan \cite{Ngo2014} predict the quiescent periods for landing a helicopter on a ship based on the model of the vessel, but they do not tackle the problem of predictions of the motion for landing on an untilted deck. This leads to short opportunistic windows that have to be adhered to, even if the conditions change rapidly. All of the above-mentioned works present results only in simulation environments that are not harsh or extreme.

The research on solutions for multi-rotor aerial vehicles landing on marine vessels is recent. One of the first works by Polvara et al. \cite{Polvara2018} uses a fiducial marker located on the platform and an extended Kalman filter (EKF) that estimates the position of the USV. In contrast, the approach presented by Abujoub et al. \cite{Abujoub2019} relies on a LiDAR onboard the \ac{uav} to find the pose of the landing pad to learn the behavior of the platform by hovering above it. However, they classify the window of landing into go or no-go intervals. Both preliminary works were validated in lenient simulation conditions. 

More recently, researchers have begun testing their approaches through real-world experiments. For example, Xu et al. \cite{9188979} use a fiducial marker for a decentralized approach, so as to follow the \ac{usv} and use a PD controller for landing once the \ac{usv} is discovered. For the second challenge of achieving decentralization, Lee et al. \cite{Lee2021} present an interesting solution to finding a ship and its pose using classical vision algorithms. Zhang et al. \cite{Zhang2021} take a different approach and present a learning-based linear controller that receives inputs from a fiducial marker to land the \ac{uav} on a \ac{usv} that is subject to the waves of a lake. Furthermore, some works also present the application of an MPC controller that enables a flexible-blade helicopter to land on a marine vessel \cite{Sultan2020, Sultan2022}. These works use a non-linear \ac{mpc} to achieve near-perfect performance but do so using a numerical benchmark that doesn't run in real-time or in a real-world experiment. Our work differs from these by using simplifications and a new approach that fills these gaps of real-time computing and applicability without a significant drop in landing performance. We use these for comparison in our experimental section to demonstrate the same. The most advanced research presented with real-world flight data is the work by Persson et al. \cite{Persson2019}, which presents an \ac{mpc} for a \ac{uav} autonomous landing on a moving boat. 

For the purpose of our work, we assume that the \ac{usv} can be found by the \ac{uav} by ascending to a given altitude during the mission without the need for conducting a planned search which is beyond the scope of this paper. We also assume that the motion of the \ac{usv} perpendicular to the water surface is minimal (the \ac{usv} is waiting for the \ac{uav} to land while controlling its global positioning on the water in order to remain stationary, rather than drifting with the waves). Furthermore, we assume that the \ac{usv} is under the influence of waves, which results in periodic oscillations of the \ac{usv} deck in each axis of a coordinated system with an origin at the \ac{usv} center of mass. For hardware, we assume that the \ac{uav} is equipped with a 2MP downward facing camera, an onboard computer for image processing and computing the \ac{mpc}, and that the \ac{usv} is equipped with a landing pattern to recognize relative pose.

The main difference between our proposed approach and \cite{Persson2019} is that our controller uses the non-linear model of the \ac{usv} for landing on a rapidly tilting deck and does not employ any communication between the \ac{uav} and the \ac{usv}, as motivated by real-world applicability. To the best of the authors' knowledge, it is the first approach using \ac{usv} motion prediction in control feedback of a decentralized controller. In summary, our contributions are: (1) we present a novel objective function for finding an optimum landing trajectory that utilizes an \ac{mpc} algorithm to predict the future of the \ac{uav} and \ac{usv}, without communication; (2) we propose a decentralized vision-based method for observing and predicting the motion of a \ac{usv} through the use of an online observer that adapts the \ac{usv} motion model using observations from a downward-facing camera; (3) our proposed approach enables landing on a highly undulating platform with no prior knowledge of the dimensions of the \ac{usv}; and (4) we propose a prediction algorithm that is designed to prevent a velocity overshoot at the set point for landing with minimal impulse transfer from the surface upon touchdown. 
The relevant media from this work has been made available as supplementary material on \href{\medialink}{\medialink}.

\begin{figure*}[!t]
    \centering
    \input{NEW_diagram.tex}
    \caption{
      The \emph{\ac{mpc}} landing controller (yellow block) is integrated into the MRS system (grey blocks) and supplies the desired reference (velocity $\mathbf{\dot{r}}_d=\begin{bmatrix} \dot{x} & \dot{y} & \dot{z}\end{bmatrix}^T$ and heading rate $\dot{\eta}_d$). In the MRS system, the first layer containing a \emph{Reference tracker} processes the desired reference and gives a full-state reference $\bm{\chi}$ to the attitude controller. The feedback \emph{Position/Attitude controller} produces the desired thrust and angular velocities ($T_d$, $\bm{\omega}_d$) for the Pixhawk flight controller (Attitude rate controller). The \emph{State estimator} fuses data from \emph{Odometry \& localization} methods to create an estimate of the \ac{uav} translation and rotation ($\mathbf{x}$, $\mathbf{R}$). Finally, the \emph{Vision-based Detector} obtains the visual data from the camera and sends the pose information \textbf{b} of the \ac{usv} to the \ac{mpc}.
      \label{fig:new_pipeline}
    }
\end{figure*}

\section{Proposed Non-linear Estimator-based MPC}\label{sec:proposed_pipeline}

In this section, we present our proposed approach which consists of a \ac{uav} prediction model and a simplified \ac{usv} prediction model. Our proposed controller must satisfy two hard constraints imposed by real-world conditions, which are: 1) The controller must perform its computation under a time constraint of 50 ms (20 Hz); and 2) There is no communication between the UAV and the USV and so, the only method for estimating the state of the USV motion is by visual pose estimation enabled by the AprilTag on the landing platform. 
Thus, for the sake of clarity, we will call our approach \textbf{\ac{mpc}-NE} (Model Predictive Controler - Non-linear Estimator).
Figure \ref{fig:new_pipeline} presents the control pipeline used in this work; the contribution is encapsulated in Figure \ref{fig:new_pipeline}. For the \ac{uav} prediction model, a discrete linear time-invariant system is used, while the \ac{usv} model uses a more complex linearised model to be described subsequently. The 6-\ac{dof} \ac{usv} pose $\mathbf{\usvaxis} = \begin{bmatrix} \usvaxis_1 & \usvaxis_2 &  \usvaxis_3 & \usvaxis_4 & \usvaxis_5 & \usvaxis_6 \end{bmatrix}^T$ is estimated in the world frame through the detection of the fiducial tag in the center of the landing pad from the on-board camera of the \ac{uav}. The pose of the \ac{uav} is fused and accounted for to estimate the correct world frame pose of the \ac{usv}. This pose information is fed to a \ac{fft} node (based on \cite{FFTW05}) which identifies the frequencies, amplitudes, and phases of the $N$ periodic oscillations that make up the \ac{usv} motion in pitch and roll axes. These identified modes are used to initialize a linear Kalman observer node that corrects the observed state and predicts future motion. These predictions are sent to the \ac{mpc} controller, which uses them to estimate the feasibility of landing in the near future, i.e., if a sufficiently low tilt of the \ac{usv} can be found inside the predefined prediction horizon. In turn, it generates the desired linear velocities for $x$, $y$, and $z$ axes, as well as the desired angular velocity in heading $\heading$. The \ac{mpc} also receives the estimated \ac{uav} state vector $\dronestatevector = \begin{bmatrix} x & \dot{x} & \ddot{x} & y & \dot{y} & \ddot{y} & z & \dot{z} & \ddot{z} & \heading & \dot{\heading} & \ddot{\heading} \end{bmatrix}^T$ using onboard state estimation proposed by our team in \cite{baca2021mrs}.

A finite-state automaton-based approach is used to direct our mission. Based on this, a setpoint generator node commands the aircraft to increase its altitude until the vision marker can be found. Once it is found, the reference of the \ac{mpc} is changed by the setpoint generator, such that it can hover at a preset altitude above the identified marker. Subsequently, the \ac{uav} waits for enough data to be gathered so that the \ac{fft} accuracy threshold requirement can be met. Once it is met, the setpoint generator sets the global reference for landing. Then, the \ac{mpc} begins to use the future motion predictions of the wave to determine a suitable time for landing.

\subsection{USV Prediction Model}

\ac{usv} models can be classified into two different types: Maneuvering Theory and Seakeeping Theory \cite{fossen2002marine}.

Owing to the assumptions made in Section \ref{related_works}, we choose to focus on the Seakeeping theory since it concerns near-stationary vessels. In addition, the use of a decentralized approach brings challenges in estimating the true odometry of the \ac{usv}, as converging to reliable estimates of linear and angular velocities of the \ac{usv} is infeasible. Therefore, we leverage the pose estimate from the camera efficiently by focusing only on the kinematics of the \ac{usv}.

Our \ac{usv} prediction model is composed of three parts: a fast Fourier transform, a Kalman observer, and a wave prediction model. First, the \ac{fft} performs a decomposition on the pose data obtained from the vision pipeline. The identified modes of these oscillations are used to initialize a Kalman observer that adapts the amplitude and the phase of the wave using the observed values online. Finally, the amplitudes and phases from the Kalman observer are sent to the wave prediction model to enable future wave predictions.

\subsubsection{FFT-based Modelling}
We assume that the motion of the \ac{usv} is composed of $N_j$ periodic waves and a non-periodic term that accounts for random noise in tracking the various components for each $j^{th}$ axis. Thus, let the state vector $\mathbf{\usvaxis}$ be represented by the linear pose $\usvaxis_j$ for $j \in \{1,2,3\}$ for x, y, z axes, respectively, and the angular pose be represented by $j \in \{4,5,6\}$ about these axes in the same order. 

Note here that, a sufficiently large ship/boat (intended application) would exhibit sufficiently low amplitude oscillations in Z-axis such that they can be handled by changing the reference at every camera frame (as shown \href{\medialink}{here}). Thus, the periodic motion of the \ac{usv} in an axis can be represented as a function of time such that:

\begin{equation}
\begin{aligned}
     \usvaxis_j(t) &= \usvaxis_{j,off} + \sum_{i=1}^{N_j}\ampji \sin{\underbrace{(2\pi \freqji t + \phaseji)}_{\Phase_{j,i}}},
\end{aligned}
\end{equation}
\noindent with $\freqji$ denoting the frequency, $\ampji$ the amplitude, and $\phaseji$ the phase. Additionally, $\usvaxis_{j,off}$ is the non-periodic term accounting for random noise. For the initial condition, $\Phase_{j,i}(t)$ is equal to $\Phase_{j,i}(\identime)$, which is the phase obtained as the output of \ac{fft} at the time of identification $\identime$. 
In sea conditions, these frequency components can change frequently due to changing winds. Therefore, the pose is sampled continuously and an \ac{fft} is run every $\fftdelta$ seconds. For each axis, we discard the modes that are below a certain threshold amplitude $\amp_{j,threshold}$, where
\begin{equation}
    \amp_{j,threshold} = \amp_{gate} \cdot \max\{\amp_{j,0}, \amp_{j,1}, \ldots, \amp_{j,N_j}\}.
\end{equation}

For reliable performance, and upon tuning on real-world data, we assume \(\amp_{gate} (=0.02)\) to be a suitable cutoff. This prevents us from identifying noise components as low-amplitude periodic oscillations without losing more than 2\% of the accuracy. These erroneous components cause a loss of performance in the Kalman observer, which is explained in the next section. 

\subsubsection{Kalman Observer}
The Kalman observer uses a linear model to refine the estimate of identified amplitude and phase of each mode. The observer is necessary because, while the \ac{fft} accurately identifies the frequency components, the amplitude and phase outputs are averages for the entire $\fftdelta$ sampling interval. Therefore, the observer receives new parameters for all identified modes every $\fftdelta$ seconds. In order to allow sufficient time for the observer parameters to converge to true values, we do not reinitialize the pre-identified modes with the new parameters. Instead, only the newly identified modes are initialized, while discarding the old modes that no longer exist.

To assemble the model, we first write the \ac{ode} for \textit{each mode} for a given axis at any time $t$. We use $\usvstatevect_{j,i}$ to denote the $i^{th}$ mode of the \ac{usv} state vector component $\usvaxis_j$ in the $j^{th}$ axis. Thus, the derivative of the mode ($\forall j, j \in {1\ldots 6}$) is
\begin{small}
\begin{equation}
    \usvstatevectder_{j,i} = \underbrace{\begin{bmatrix}
    0 & 1\\
    -(2\pi\freqji)^2 & 0\\
    \end{bmatrix}}_{\singlestatematrix(\identime)} \usvstatevect_{j,i},
\end{equation}
\end{small}

\noindent and the mode at time $t$ is
\begin{small}
\begin{equation} \label{eq:statevector}
    \usvstatevect_{j,i} = \begin{bmatrix}
    \ampji \sin{(\Phase_{j,i}(t))} \\
    2\pi\ampji\freqji\cos{(\Phase_{j,i}(t))}
    \end{bmatrix}.
\end{equation}
\end{small}

Next, we derive the observer model by adding the \acp{ode} of each mode. Thus,
\begin{small}
\begin{equation}
    \usvstatevectder_j(t) = \underbrace{\begin{bmatrix}
    \singlestatematrix_{j,1} & \mathbf{0} & \ldots & \mathbf{0} & \mathbf{0} \\
    \mathbf{0} & \singlestatematrix_{j,2} & \ldots & \mathbf{0} & \mathbf{0} \\
    \vdots & \vdots & \ddots & \vdots & \vdots \\
    \mathbf{0} & \mathbf{0} & \ldots & \singlestatematrix_{j,N} & \mathbf{0}\\
    \mathbf{0} & \mathbf{0} & \cdots & \mathbf{0} & \mathbf{0}\\
    \end{bmatrix}}_{\combinedstatematrix_j}
    \underbrace{
    \begin{bmatrix}
    \usvstatevect_{j,1} \\
    \usvstatevect_{j,2} \\
    \vdots \\
    \usvstatevect_{j,N_j} \\
    \usvstatevect_{j,off} \\
    \end{bmatrix}}_{\usvstatevect_j(t)}.
\end{equation}
\end{small}

Hence, the output for each axis is
\begin{small}
\begin{equation}
\begin{aligned}
    \usvaxis_j(t) = \underbrace{\begin{bmatrix}
    \singleoutputmatrix_{j,1} & \singleoutputmatrix_{j,2} & \cdots & \singleoutputmatrix_{j,N} & \singleoutputmatrix_{j,off}\\
    \end{bmatrix}}_{\combinedoutputmatrix_j}\usvstatevect_j(t).
    \end{aligned}
\end{equation}
\end{small}

Note, that each component of the output vector of the mode can be found as
\begin{equation}
    \usvaxis_{j,i} = \underbrace{\begin{bmatrix}
    1 & 0\\
    \end{bmatrix}}_{\singleoutputmatrix_{j,i}} \usvstatevect_{j,i}.
\end{equation}

Now, for the brevity of explanation and the readability of the equations, we write the following relation \textit{for only one axial \ac{dof}} of the \ac{usv} in discrete-time. In addition, we clarify that it can be applied to all 6 of the \ac{dof}. Furthermore, notice that a time instance $t = k \discsamplingtime + \identime$, wherein $\discsamplingtime$ is the discrete sampling time for new pose observations. Thus, we have a straightforward change in notation such that, for example, $\usvstatevect_j(t) \equiv \usvstatevect_{j}^{(k)}$. Thus, by using the integral approximation method, we have that
\begin{small}
\begin{equation}
\begin{aligned}
\discusvstatevect_j^{(k+1)} &= \underbrace{\exp({\combinedstatematrix_j \discsamplingtime})}_{\discstatematrix_j} \discusvstatevect_j^{(k)},\; and \quad
\discusvaxis_j^{(k)} &= \combinedoutputmatrix_j \hspace{2pt}\discusvstatevect_j^{(k)}.
\end{aligned}
\end{equation}
\end{small}

Then, we continuously estimate the amplitude $\ampji$ and phase $\phaseji$ of each mode every $\discsamplingtime$ using the Kalman Filter. First, $\systemnoisemat$ is initialised using a diagonal matrix $\systemnoisematI = \lambda\mathbf{I}$, such that
\begin{small}
\begin{equation}
    \systemnoisemat = \frac{1}{2} (\discstatematrix  \systemnoisematI  \discstatematrix^T + \systemnoisematI)  \discsamplingtime,
\end{equation}
\end{small}\noindent
where $\lambda$ is the gain parameter for the process noise observed in the model.

Meanwhile, the observation noise matrix $\observationnoise$ is set to the mean amplitude of the observed noise in the system. Thereafter, we apply the filter equations as follows:

\begin{small}
\begin{equation}
\begin{aligned}
    \hat{\discusvstatevect}_j^{(k)} &= \discstatematrix_j \discusvstatevect_j^{(k-1)},\\
    \hat{\processcovariancemat}^{(k)} &= \discstatematrix_j\processcovariancemat^{(k-1)}\discstatematrix_j^T + \systemnoisemat,\\
    \hat{\discusvaxis}_j^{(k)} &= \combinedoutputmatrix_j\hat{\discusvstatevect}_j^{(k)},\\
    \kalmangain^{(k)} &= \hat{\processcovariancemat}^{(k)}\combinedoutputmatrix_j^T(\combinedoutputmatrix_j\hat{\processcovariancemat}^{(k)}\combinedoutputmatrix_j^T + \observationnoise)^{-1},\\
    \usvstatevect_j^{(k)} &= \hat{\usvstatevect}_j^{(k)} + \kalmangain^{(k)}(\discusvaxis_{j,m} - \hat{\discusvaxis}_j^{(k)}),\\
    \processcovariancemat^{(k)} &= (\mathbf{I} - \kalmangain^{(k)}\combinedoutputmatrix_j)\hat{\processcovariancemat}^{(k)},\\
\end{aligned}
\end{equation}
\end{small}\noindent
where $\hat{}$ shows the predicted value of the vector/matrix, $\mathbf{I}$ is an identity matrix, $\discusvaxis_{j,m}$ is the measured value of $\discusvaxis_j$, $\kalmangain \in \mathbb{R}^{2(N+1)}$ is the Kalman gain matrix of the system, $\processcovariancemat$ and $\systemnoisemat \in \mathbb{R}^{2(N+1) \times 2(N+1)}$ are the process co-variance and system noise matrices, respectively, and $\observationnoise \in \mathbb{R}$ is observation noise. 

At every identification, the relevant elements corresponding to both of the modes that are no longer present, as well as the newly identified modes of the $\discstatematrix$ matrix, are re-initialized. The corresponding co-variance terms for these modes are reset to maintain a consistent prediction without large deviations.

\subsubsection{Wave prediction}
Let us now define $t_{obs}$ as the time instant where the last observation was performed, since the prediction algorithm is not run when there are no new observations. Thus, by running the Kalman observer at $t_{obs}$ we find the new amplitude $\obsAmp$ and phase $\obsPhase$. At the same instant in time $t_{obs}$, we can extract the corresponding $\usvstatevect_{j,i}$ and use (\ref{eq:statevector}) to acquire:
\begin{small}
\begin{equation}
    \begin{aligned}
    \obsPhase &= \arctan \left(\frac{2\pi\freqji[\usvstatevect_{j,i}]^{1,1}}{[\usvstatevect_{j,i}]^{2,1}}\right),\\
    \text{and}\\
    \obsAmp &= \frac{[\usvstatevect_{j,i}]^{1,1}}{\sin{(\obsPhase)}},\\
    \end{aligned}
\end{equation}
\end{small}
where $[\usvstatevect_{j,i}]^{m,n}$ represents the element corresponding to the $m^{th}$ row and $n^{th}$ column of the vector.

This enables us to predict the wave behavior at a future time $t > t_{obs}$ as
\begin{small}
\begin{multline}
    \label{eq:usv_model}
    \usvaxis_j(t) = \sum_{i=1}^{N_j}\obsAmp \sin{\left[2\pi \freqji (t - t_{obs}) + \obsPhase\right]} \\
     + [\usvstatevect_{j,off}]^{1,1}.
\end{multline}
\end{small}

\subsection{UAV Prediction Model}

The \ac{uav} prediction model used in the proposed \ac{mpc} is based on the Euler approximation of a set of single particle kinematics equations. Here, we employ the following discrete linear time-invariant system:
\begin{footnotesize}
\begin{equation}
    \dronestatevector^{(k+1)} = \dronestatematrix\dronestatevector^{(k)} + \droneinputmatrix\droneinputvector^{(k)},    \label{eq:uav_state_eqn}\; with \quad
    \droneinputvector^{(k)} =
    \begin{bmatrix}
        \dot{\ddot{x}} & \dot{\ddot{y}} & \dot{\ddot{z}} & \dot{\ddot{\heading}} 
    \end{bmatrix}^T.
\end{equation}
\end{footnotesize}

In the model represented by (\ref{eq:uav_state_eqn}), the state matrix $\dronestatematrix$ and the input matrix $\droneinputmatrix$ can be found through the Kronecker product ($\otimes$), such that:
\begin{small}
\begin{equation}
    \underset{12\times12}{\dronestatematrix} =
    \underset{4\times4}{\mathbf{I}} \otimes \underset{3\times3}{\substatematrix},\; with \quad
    \substatematrix = \begin{bmatrix}
    1 & \deltapred & \frac{\deltapred^2}{2}\\
    0 & 1 & \deltapred\\
    0 & 0 & 1
    \end{bmatrix},
\end{equation}
\end{small}

\begin{small}
\begin{equation}
   \underset{12\times4}{\droneinputmatrix} = \underset{4\times4}{\mathbf{I}} \otimes \underset{3\times1}{\subinputmatrix},\; with \quad
   \subinputmatrix = 
   \begin{bmatrix}
      \frac{\deltapred^3}{6} \\
      \frac{\deltapred^2}{2} \\
      \deltapred 
   \end{bmatrix},
\end{equation}
\end{small}

\noindent where $\mathbf{I}$ is an identity matrix, with a prediction made every $\deltapred = 0.01$ seconds.

Hence, the state vector represents the states of the system and their derivatives up to acceleration in each axis, and the control input is the jerk experienced in those axes.

\subsection{MPC Objective Function}

Once we have defined a prediction model of the \ac{uav} and the \ac{usv}, we can formulate an objective function to enable both way-point navigation and landing. For the sake of simplification, we will omit the superscript $(.)^{(k)}$, which represents a discrete instant in time. Therefore, we can define the objective function $J$ as:

\begin{small}
\begin{equation}
\begin{aligned}
    \min_{\droneinputvector_1,\ldots,\droneinputvector_{\controlhorizon}} J(\dronestatevector, \droneinputvector) &= \underbrace{\sum_{m = 1}^{\predhorizon}\errormat_m^T \errpenmat \errormat_m + \inputeffortvect_m^T \inputeffortpen \inputeffortvect_m}_{J_1}\\ &+ \underbrace{\sum_{m = 1}^{\predhorizon}\alpha_L \times g(\errinz_m, \usvaxis_{4,m}, \usvaxis_{5,m})}_{J_2},\\
    \text{subject to :}\\
    \errormat_m &= \dronestatevector_m - \dronestatevectdes_m,\\
    \Tilde{z}_m &= z_m-\overset{*}{z}_m,\\
    \inputeffortvect_m &= \droneinputvector_m-\droneinputvector_{m-1},\\
    \dronestatevector_{m+1} &= \dronestatematrix\dronestatevector_m + \droneinputmatrix\droneinputvector_m~\forall~m \leq \controlhorizon, \\
    \dronestatevector_{m+1} &= \dronestatematrix\dronestatevector_m + \droneinputmatrix\droneinputvector_{\controlhorizon}~\forall~m > \controlhorizon, \\
    \droneinputvector_{min} &\leq \droneinputvector_m \leq \droneinputvector_{max},\\
    \dronestatevector_0 &= \dronestatevector_{initial},\\
    \droneinputvector_0 &= \droneinputvector_{initial},\\
    &\forall~\{m: m\in\mathbb{N}, 1 \leq m \leq \predhorizon\},
\end{aligned}
\end{equation}
\end{small}
where $\dronestatevectdes_m$ is the desired state, $\errormat_m$ is the error vector, $\errinz_m$ is the error in $z_m$ position,  $\inputeffortvect_m$ is the rate of control input change to ensure smooth input to the \ac{uav}, $\predhorizon(=100)$ is the prediction horizon, and $\controlhorizon(=40)$ is the control horizon. $\errpenmat$ and $\inputeffortpen$ are the corresponding penalty matrices with configurable weights for performance tuning, while $\alpha_L(=1200)$ is a weight chosen for the tuning of the objective function $g(.)$. Additionally, $\usvaxis_{4,m}, \usvaxis_{5,m}$ are the pitch and roll angles in discrete time of the \ac{usv} about its $x$ and $y$ axes, according to (\ref{eq:usv_model}).

We emphasize that $\dronestatevectdes_m$ (including $\overset{*}{z}$) can either be a series of points (trajectory) or a single point (step input). This would enable the \ac{uav} to keep up with a drifting \ac{usv} if the XY-position state of the usv is estimated independently. However, a slowly drifting \ac{usv} is within the dynamic limits of the \ac{uav} so as to be compensated by the single-point reference that can be updated after every observation (depending on the camera frame rate). We demonstrate and test this in \href{\medialink}{this linked video}.
While we do not constrain the output of the \ac{mpc}, we apply soft constraints to the velocity and acceleration states of the model, such that $v \geq v_{max} \text{ and } a \geq a_{max}$ incur a high penalty in the objective function. 

Herein, we have introduced a novel objective function $J_2$ (described in the next section) which can account for the predicted motion of the \ac{usv}, producing a smooth control input to change the altitude of the \ac{uav} without any abrupt maneuvers. Using this function, we are able to incorporate the finite state automaton approach using a sigmoid activation function without explicitly describing the possible landing condition. The \ac{uav} is able to follow the descend trajectory generated by the \ac{mpc} by autonomously adjusting its hover distance above the \ac{usv}. Additionally, it enables us to tune the parameters to control the variance of the resulting landing angles about the mean value of zero-tilt. It is important to mention that the term $J_1$ in our cost function, is a classical quadratic objective function largely used in robotics and well documented for its feasibility and stability. On the other hand, the $J_2$ term is different from usual works in the literature because it tackles the terminal cost of the optimization step as a potential barrier.

We employ a non-linear optimization library (NLOPT\cite{Gablonsky2001}\cite{NLOPT}) which provides the near-optimum solution for the objective function. In order to exercise velocity-based control, the first input from the series of optimum control inputs calculated by the solver is then used to calculate the next state using (\ref{eq:uav_state_eqn}). The velocities for this predicted next state are then passed to the system as the velocity references for the \ac{uav} to track (as seen in Figure~\ref{fig:new_pipeline}).

Since the term $J_1$ primarily contributes to the position control and $J_2$ contributes to the landing approach, the term $J_2$ remains disabled until the conditions for the landing approach are satisfied.

\subsection{Landing approach}

We define the function for landing cost as a combination of sigmoids, such that:
\begin{small}
\begin{equation}\label{eqn:J-2}
    g(\errinz_m, \usvaxis_{4,m}, \usvaxis_{5,m}) = f(\errinz_m)\cdot((\usvaxis_{4,m})^2 + (\usvaxis_{5,m})^2),
\end{equation}
\end{small}
\noindent where $f(\errinz_m)$ is such that

\begin{figure}[!t]
    \centering
    \includegraphics[width=3.5in]{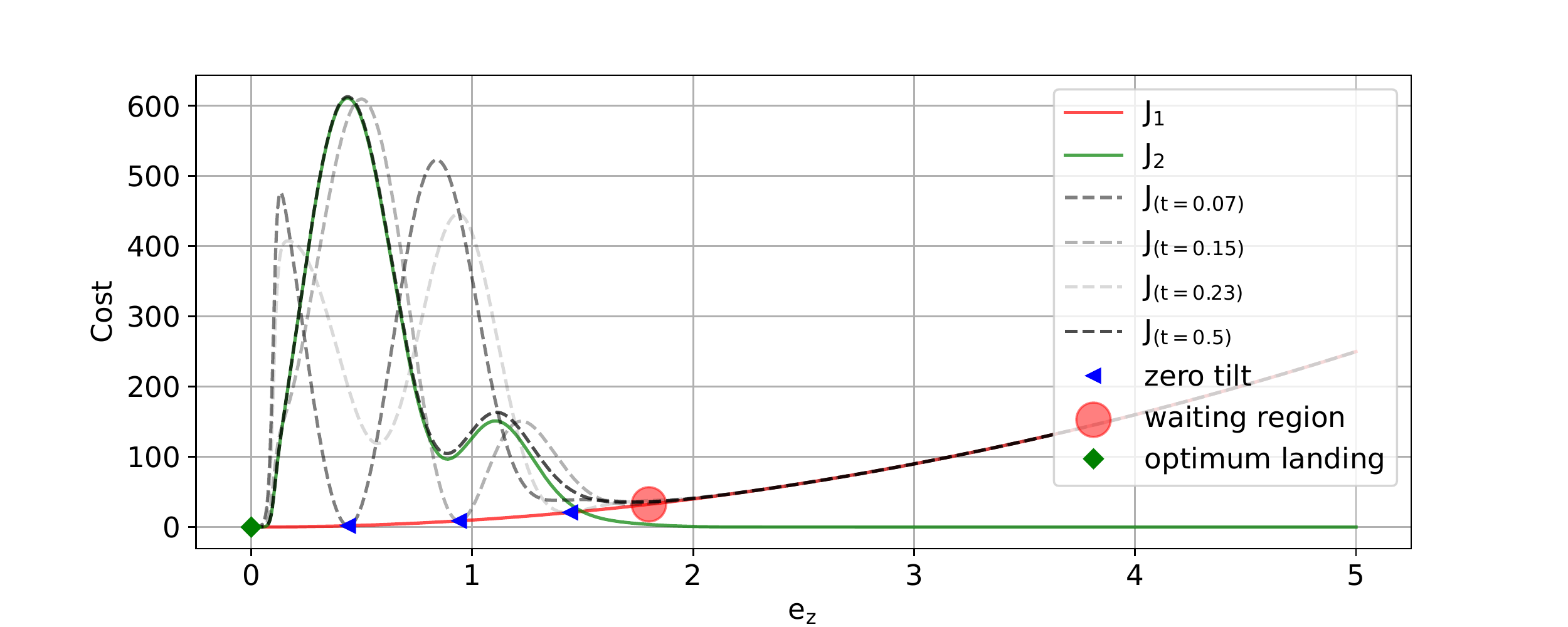}
    \caption{An example illustration of the effective cost function values obtained during the landing approach.}
    \label{fig:cost_function}
\end{figure}

\begin{footnotesize}
\begin{equation}
    f(\errinz_m)= 
\begin{cases}
    \left(1.0 + \exp\left(-\cfrac{\errinz_m - h_d}{-0.15}\right)\right)^{-1} ,& \text{if } \errinz_m \geq 0.16\\ 
    \left(1.0 + \exp\left(\cfrac{\errinz_m - 0.1}{-0.01}\right)\right)^{-1}  ,& \text{otherwise},
\end{cases} 
\end{equation}
\end{footnotesize}
\noindent where $h_d$ controls the waiting region (see Figure \ref{fig:cost_function}) during a landing attempt. Empirically, $h_d = 1.1$ was chosen for our experiments. For the scope of this paper, we assume that the \ac{usv} has relatively negligible motion in its $x$ and $y$ axes, which is a fair assumption for the problem of landing. The propulsion of the \ac{usv} may easily compensate for the drift generated by the water currents in order to facilitate landing. It is also safe to assume that $\errinz \geq 0$, as the \ac{uav} cannot approach from beneath the \ac{usv}. In order to activate $J_2$ to start the landing phase, two conditions must be met. First, \textbf{\ac{fft} accuracy is higher than a given threshold to detect slow oscillations.} Second, \textbf{The position errors in $x$ and $y$ are below a predefined threshold (i.e., $\Tilde{x}, \Tilde{y} \approx 0$) and horizontal velocities $v_x, v_y$ are also minimal.}

To demonstrate the interaction of $J_2$ with $J_1$ during the landing approach, we present a \textit{highly-simplified} plot of the objective function (see Figure \ref{fig:cost_function}) using one mode each for pitch and roll axes. 
When $J_2$ is activated, we acquire a combined plot governed by both the equation (\ref{eqn:J-2}) and the residual error $\errinz$ in $J_1$. In Figure \ref{fig:cost_function}, the value of the objective function encounters a peak that continuously evolves as a function of time. This peak acts as a potential barrier. The higher cost associated with the peak holds the aircraft in the waiting region (as marked in the plot). Meanwhile the \ac{usv} model generates predictions for the future of the \ac{usv} motion during every iteration of the \ac{mpc}. The \ac{usv} sometimes gets close enough to a zero tilt wherein a feasible solution appears, as shown by the \textit{zero-tilt} points in the plot. The \ac{uav} is then able to \textit{insert} itself into the time-varying trajectory of these special feasible points by reducing its altitude and approaching in such a way that the cost continues to decrease along the locus of these points. Therefore, the \ac{uav} is able to follow the zero-tilt points and finish at the optimum landing point, where touchdown is confirmed by the system based on thrust and other information from onboard sensors.

\section{Simulation Experiments}

We demonstrate our simulation results in two scenarios: with a numerical simulation, and with a realistic ROS-based Gazebo simulator\cite{baca2021mrs}. The \textbf{SHMPC} presented in \cite{Sultan2020} is shown to work numerically. Thus, we use a similar numerical implementation of our work (\textbf{MPC-NE}) to allow us to perform a fair comparison with the state-of-the-art. In this comparison, the non-linear optimization problem is solved by \cite{NLOPT} for a landing maneuver of 3 meters and assumes true knowledge of the future motion of the \ac{usv}. The second comparison is performed using real-time flight with our proposed \textbf{MPC-NE} inside the Gazebo simulator. For this comparison, we use a \textbf{standard MPC}\cite{baca2021mrs} designed for waypoint navigation. For this standard approach, the \ac{uav} attempts to locate the target, and lands after a programmed, uniformly randomly distributed delay between 0 and 100 seconds. We select this duration owing to the periodicity of the tilt angle of the \ac{usv}. We use a T650 quadrotor frame weighing 3.6 kg carrying a Garmin LiDAR for laser-ranging of altitude and an Intel Realsense D435 camera for live in-simulation video. The video output of the Realsense camera is sent to our system to enable processing on the vision node. (Sec. \ref{sec:proposed_pipeline}). The 3D model of the \ac{usv} is similar to our real-world experiments and is affixed with an AprilTag \cite{krogius2019iros} marker for pose estimation. We note that in \textit{both} the comparisons, we push the boundary of performance and test our work in rough sea states, and drive our \ac{usv} model using a wave generator with 4-5 components of oscillation in both pitch and roll axes, and tilt angles up to $30^\circ$($0.5$ radians). The frequency components are set such that brief windows for feasible landing exist. Wind disturbances are not considered in this environment since it is tackled by body disturbances estimated by the low-level control feedback pipeline (see Figure~\ref{fig:new_pipeline}).

\begin{figure}[t]
\centering
\subfigure{\includegraphics[width=1.6in]{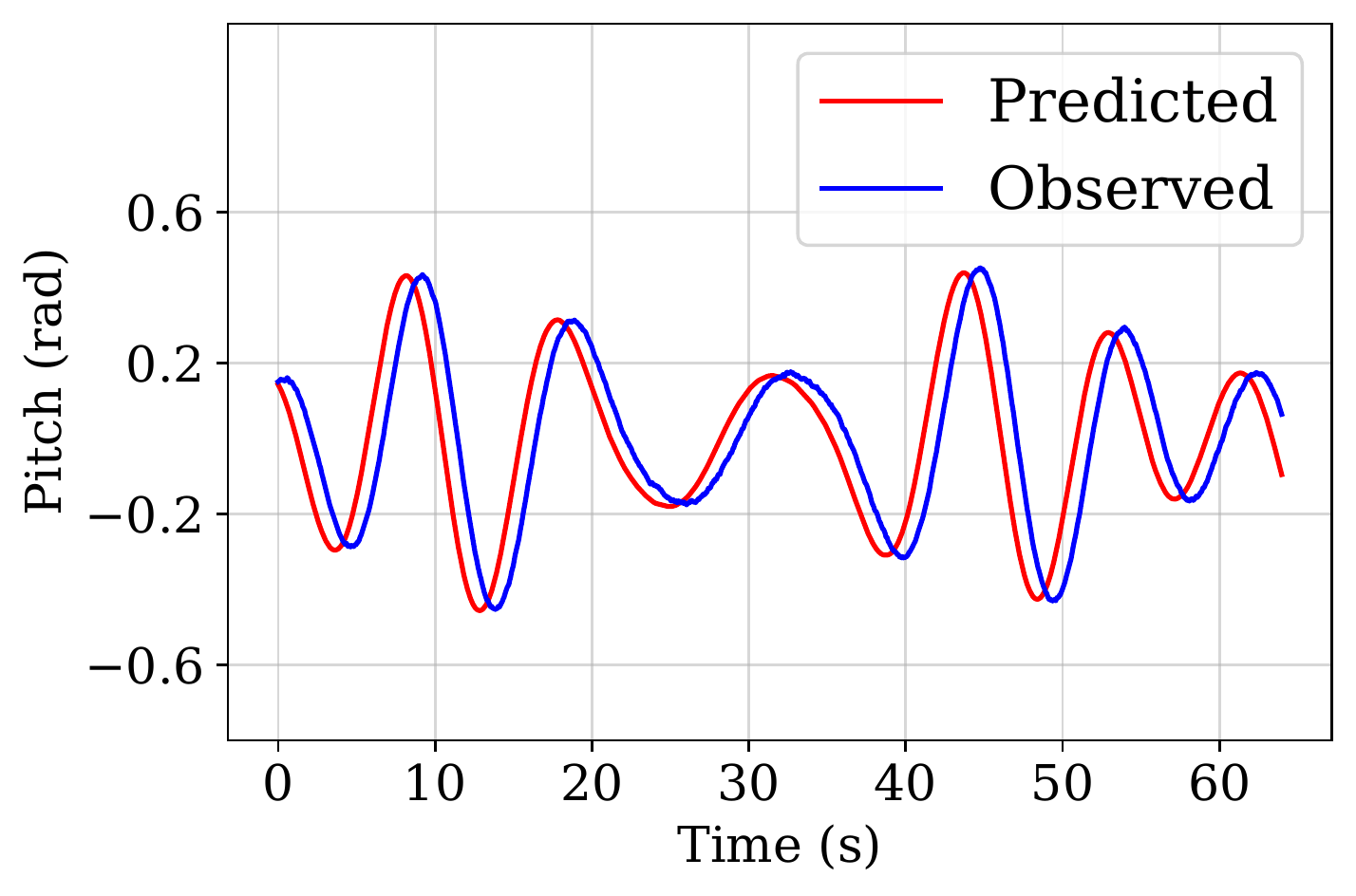}}
\hfil
\subfigure{\includegraphics[width=1.6in]{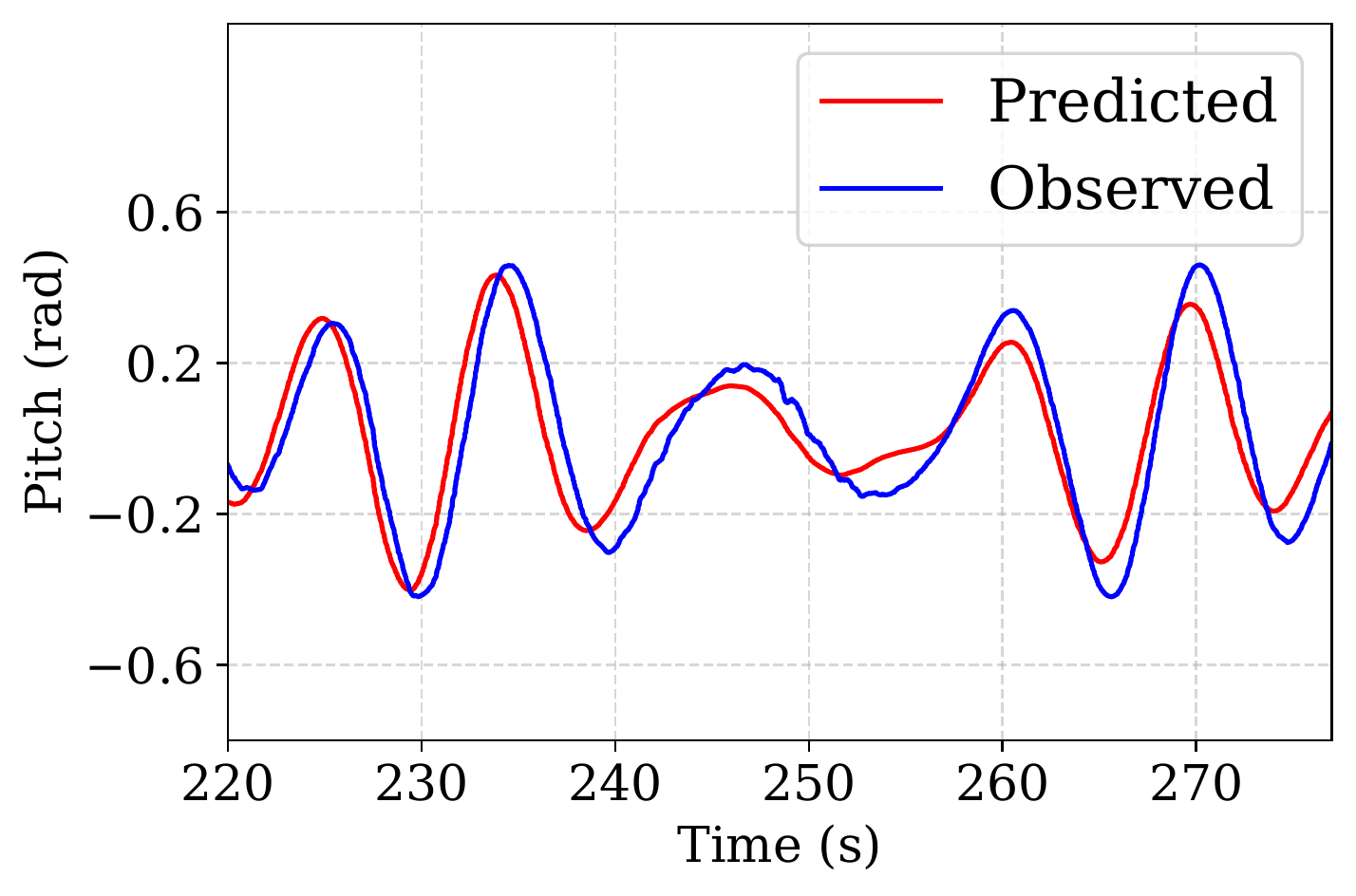}}
\caption{Comparison between the predictions made by the system using the onboard \ac{imu} data (left) and using the vision data (right).}
\label{fig:pred_comparison}
\end{figure}
\subsection{Prediction results}

First, we demonstrate the ability of our system to predict the wave motion up to 1 second into the future based on the observed frequency components and our model of the system. The performance of the system is tested in two scenarios: a 100 Hz odometry output from the simulated \ac{usv} \ac{imu} (used as ground truth), and a 30 Hz stream from the AprilTag.

As we see in Figure \ref{fig:pred_comparison}, the high-frequency \ac{imu}-based predictions are able to match the observed wave reliably without introducing noise. The observer is able to adapt the observed frequency, amplitude, and phase of the modes of the oscillations and converge reliably. As opposed to that, we see slight deviations in the vision-based predictions compared to the \ac{imu} results. This deviation in performance can be explained by two factors. First, the linearisation of the model in the time-domain causes inaccuracies that grow as the sampling time increases. Due to this, the three-fold sampling rate of the \ac{imu} leads to faster and more accurate convergence. Second, the output rate of the AprilTag identification node fluctuates around 30 Hz, depending on the computational load of the onboard computer of the \ac{uav} (or simulation computer, in this case). This leads to the misidentification of modes, as the \ac{fft} algorithm requires a fixed sample rate for observations. However, this sufficiently proves the ability of the proposed system to reliably predict wave behavior, which will be used for the \ac{usv} landing further down the pipeline.

\begin{figure}[!t]
    \centering
    \includegraphics[width=3.2in]{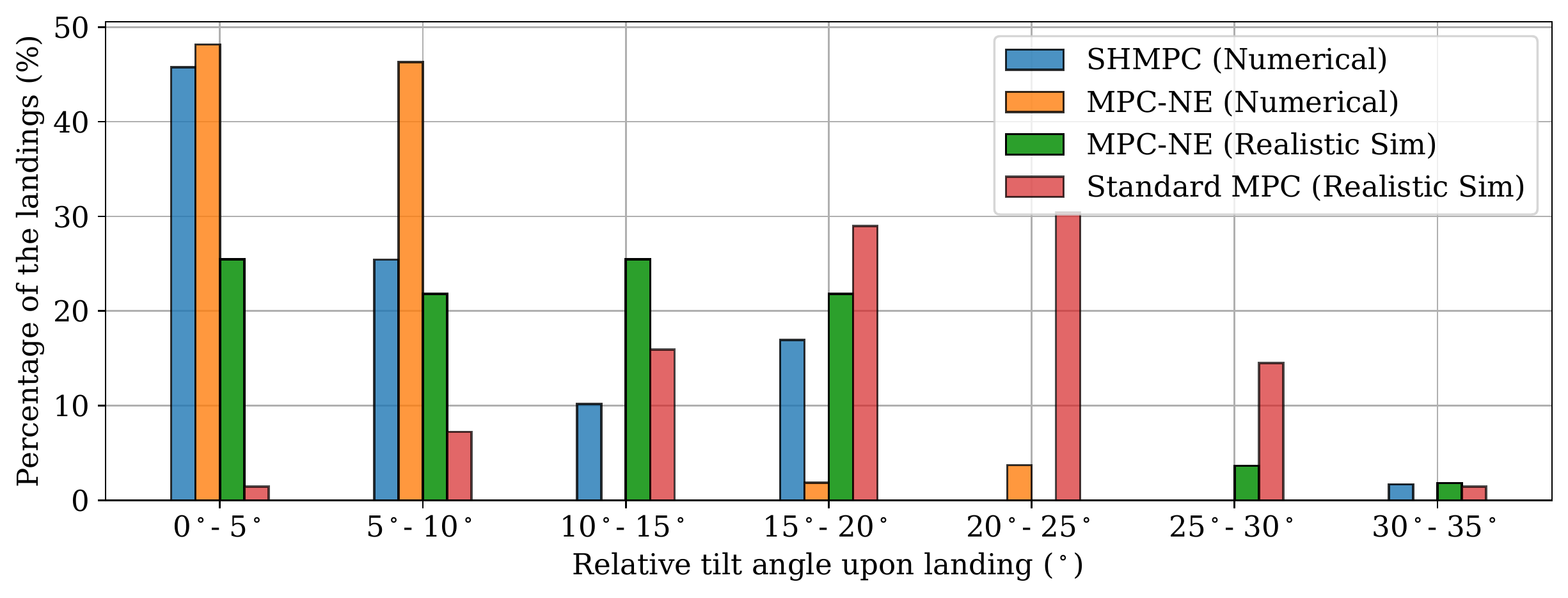}
    \caption{Histogram comparison between the proposed approach and the standard approach during the touchdown of the \ac{uav} on the \ac{usv} deck.}
    \label{fig:angle_comparison}
\end{figure}

\begin{figure*}[!b]
\centering
\subfigure[Vision---0.25 s into the future. \label{fig:025_pred}]{\includegraphics[width=2.30in]{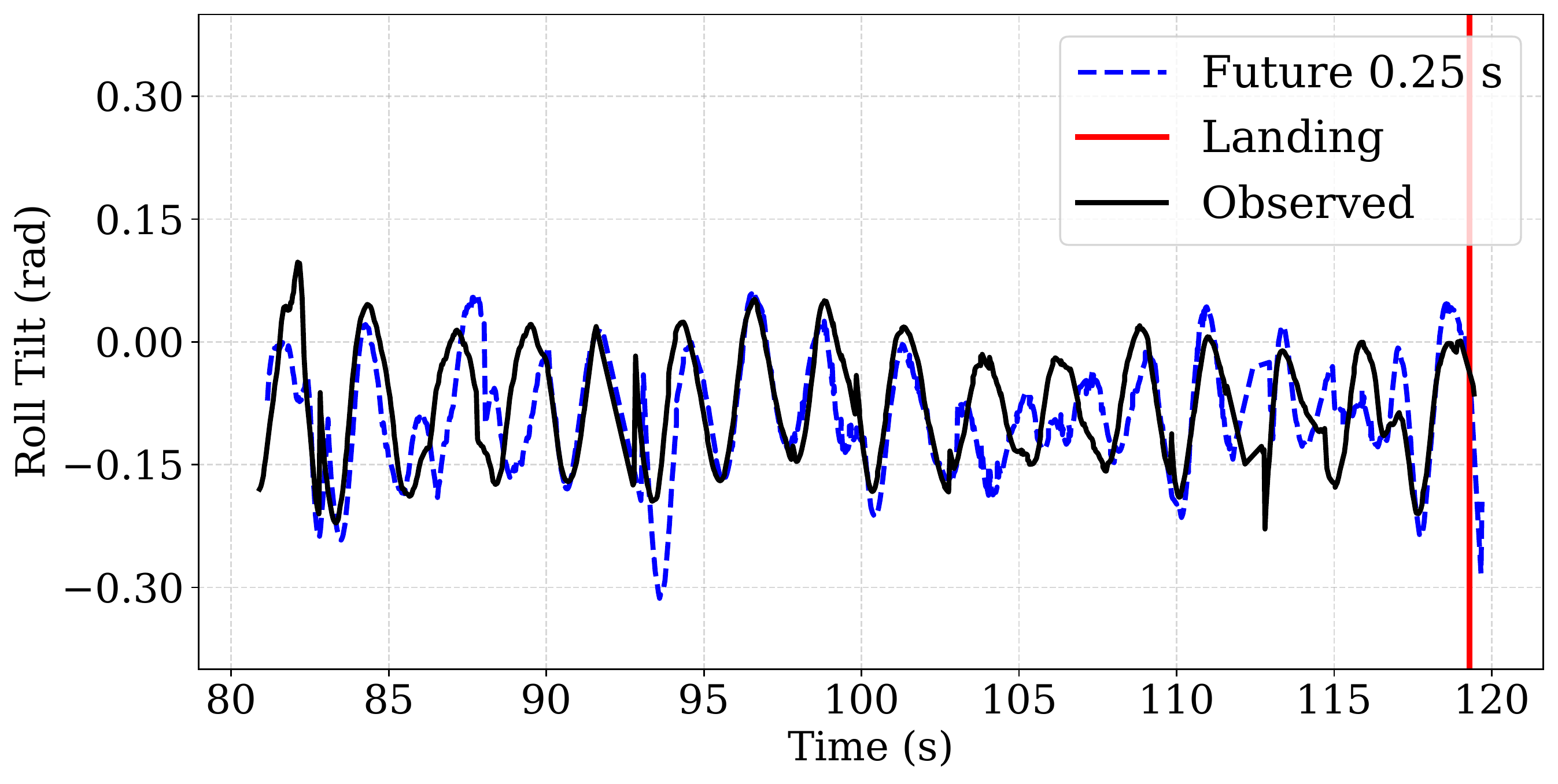}}
\subfigure[Vision---1.0 s into the future. \label{fig:100_pred}]{\includegraphics[width=2.30in]{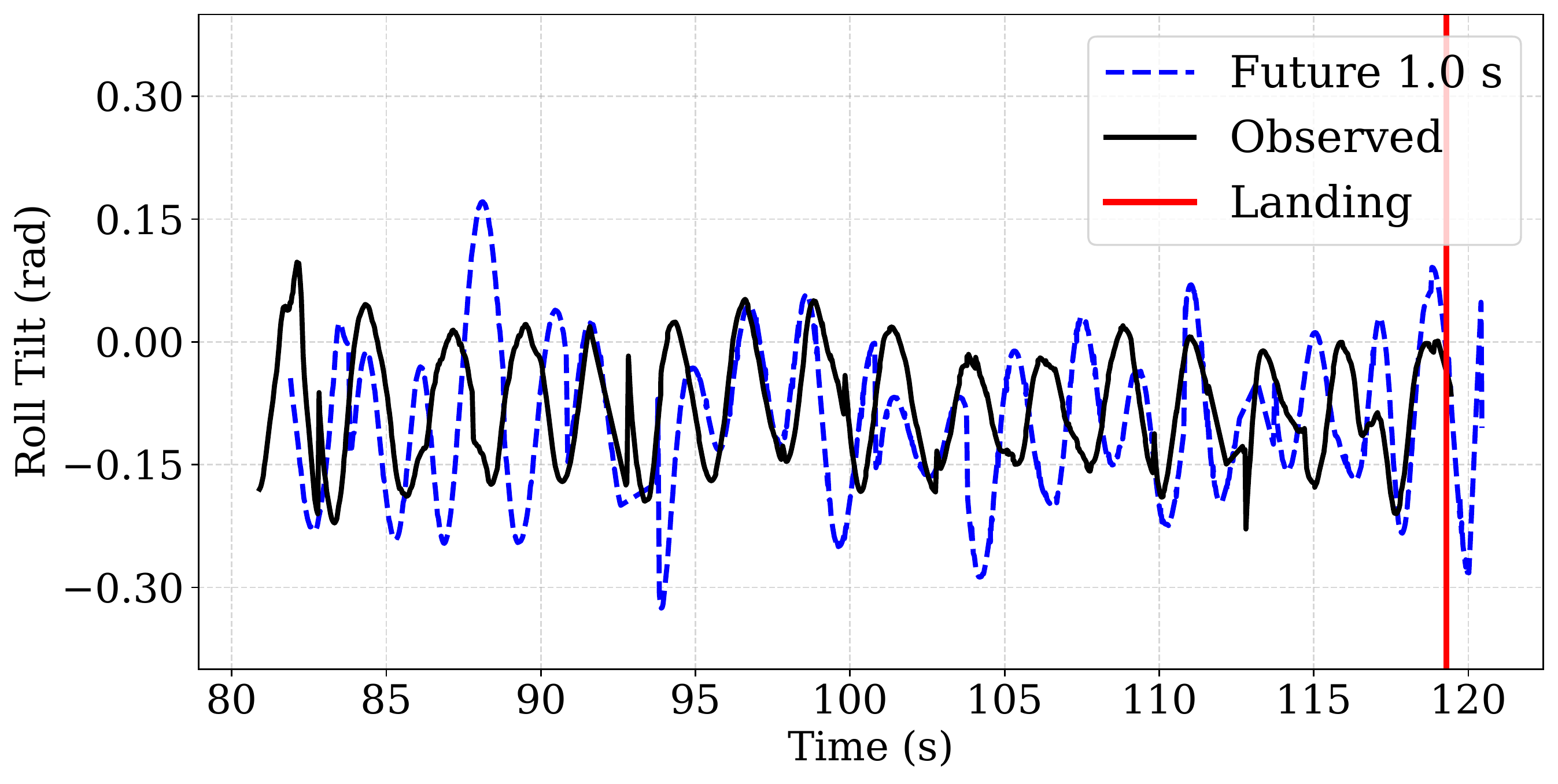}}
\subfigure[\ac{imu}---1.0 s into the future. \label{fig:imu_pred}]{\includegraphics[width=2.30in]{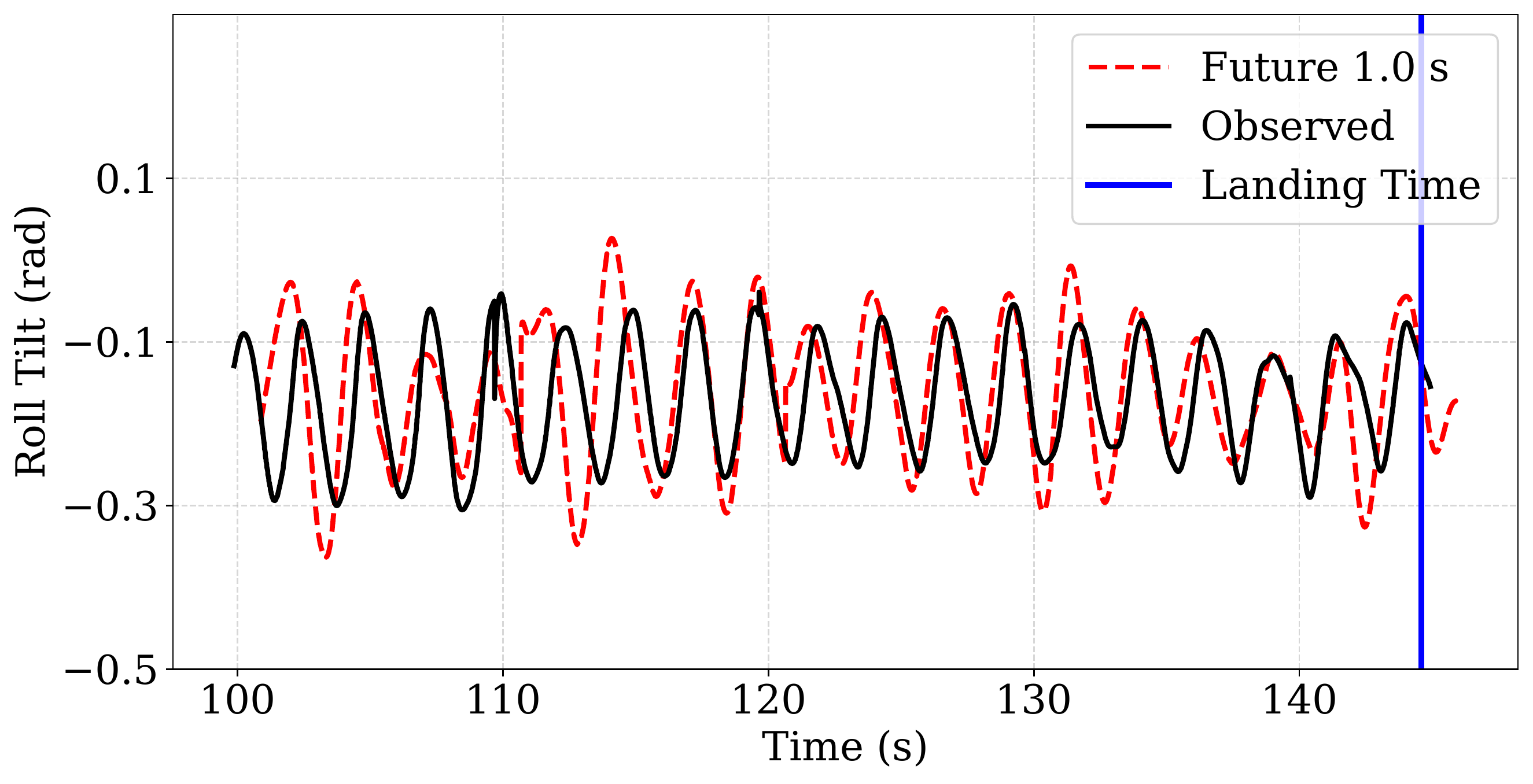}}
\caption{Comparison between the predictions made using vision (a-b) and using the onboard \ac{imu} of the \ac{usv} (c).}
\label{fig:real_pred_comparison}
\end{figure*}

\subsection{Landing results}
To continue, we present the ability of our system to land on a platform while tilt angles are sufficiently close to zero. Thus, we present Figure \ref{fig:angle_comparison} that shows the results of the numerical comparison between our MPC-NE and the state-of-the-art SHMPC. Note here that the MPC-NE lands $\approx94\%$ within $10^\circ$ of tilt, while the SHMPC lands $\approx71\%$ of the same tilt interval. In this same comparison, the solution time per iteration of our MPC-NE was 9 times lower at 102 ms compared to 917 ms for SH-MPC. 

Furthermore, the same figure presents the results of the realistic simulations using Gazebo. It is important to highlight the difference between a numerical simulation result and a realistic simulation result. This is explained by the existing constraints of processing time that demand the algorithms to be processed in real-time. Note here that the MPC-NE is able to conduct $72\%$ of its landings within $15^{\circ}$($0.26$~\text{rad}) of tilt compared to $23\%$ of landings using the standard MPC approach. In addition, the proposed approach reduces the $80^{th}$ percentile result by $9^{\circ}$($0.16$~\text{rad}) in comparison to the standard approach. For this comparison, we classify a landing conducted at a tilt angle of less than $20^{\circ}$($0.35$~\text{rad}) as successful. Therefore, even in challenging tilts of up to $0.5$ radians, the proposed approach had only three failures, while the standard approach fails approximately $50\%$ of the landings. Finally, we also highlight that, even in an unrealistic and challenging scenario, our system is able to conduct $70\%$ of the landings within $50$ seconds of reaching its \ac{fft} accuracy threshold.

\section{Real-world Experiments}

To test the contributions and proposed algorithms in the real world, we performed landings on an oscillating target at an open water reservoir. 

For the purpose of this experiment, we employed a 4.5 kg T650 quadrotor equipped with vertical pontoons\cite{9836083} for safety over water (see Figure \ref{fig:landing_image_real}). In addition, the sensor stack included a Garmin LiDAR for laser-ranging of altitude, a Basler camera for the live video feed, and an Intel NUC for onboard real-time processing of the algorithms, data, and video. The target is a special custom-made \ac{usv}\cite{9775538} equipped with a $2 m \times 2 m$ landing zone, affixed with an AprilTag\cite{krogius2019iros} for 6-\ac{dof} pose-estimation. The experimental conditions subjected the UAV to a wind of 7m/s and a \ac{usv} oscillating with an amplitude of $0.3$ radians.

\subsection{Prediction results}

Here we demonstrate our prediction pipeline in two scenarios: a 30 Hz stream from AprilTag, and a 100 Hz stream from the \ac{imu}. The prediction results for the real-world experiments are presented in Figure \ref{fig:real_pred_comparison} and discussed below.

For predictions based on vision-based pose estimation, as seen in Figure \ref{fig:025_pred}, the near-term future correlates well with the observed motion. However, Figure \ref{fig:100_pred} indicates that the predictions for the long-term future can suffer in accuracy.
This correlates well with the simulation results as shown in Figure \ref{fig:pred_comparison} and can be attributed to the higher sampling time-step and its higher variability. Occasionally, it also exhibits convergence and consecutive divergence as more data is fed into the pipeline. For ground truth, we use Figure \ref{fig:imu_pred} to demonstrate the effectiveness of the pipeline in robustly predicting the future of the \ac{usv}. However, since \ac{mpc} exhibits a higher reliance on the predictions that are temporally proximal, the predictions for $0.25$ and $0.50$ seconds into the future offer robust support for preventing a landing during an infeasible window. The chosen angle for landing is also sufficiently low in order to demonstrate the prediction capabilities and the selection of a feasible landing window. 

\subsection{Landing results}
\begin{figure}[!t]
\centering
\subfigure{\includegraphics[width=3in]{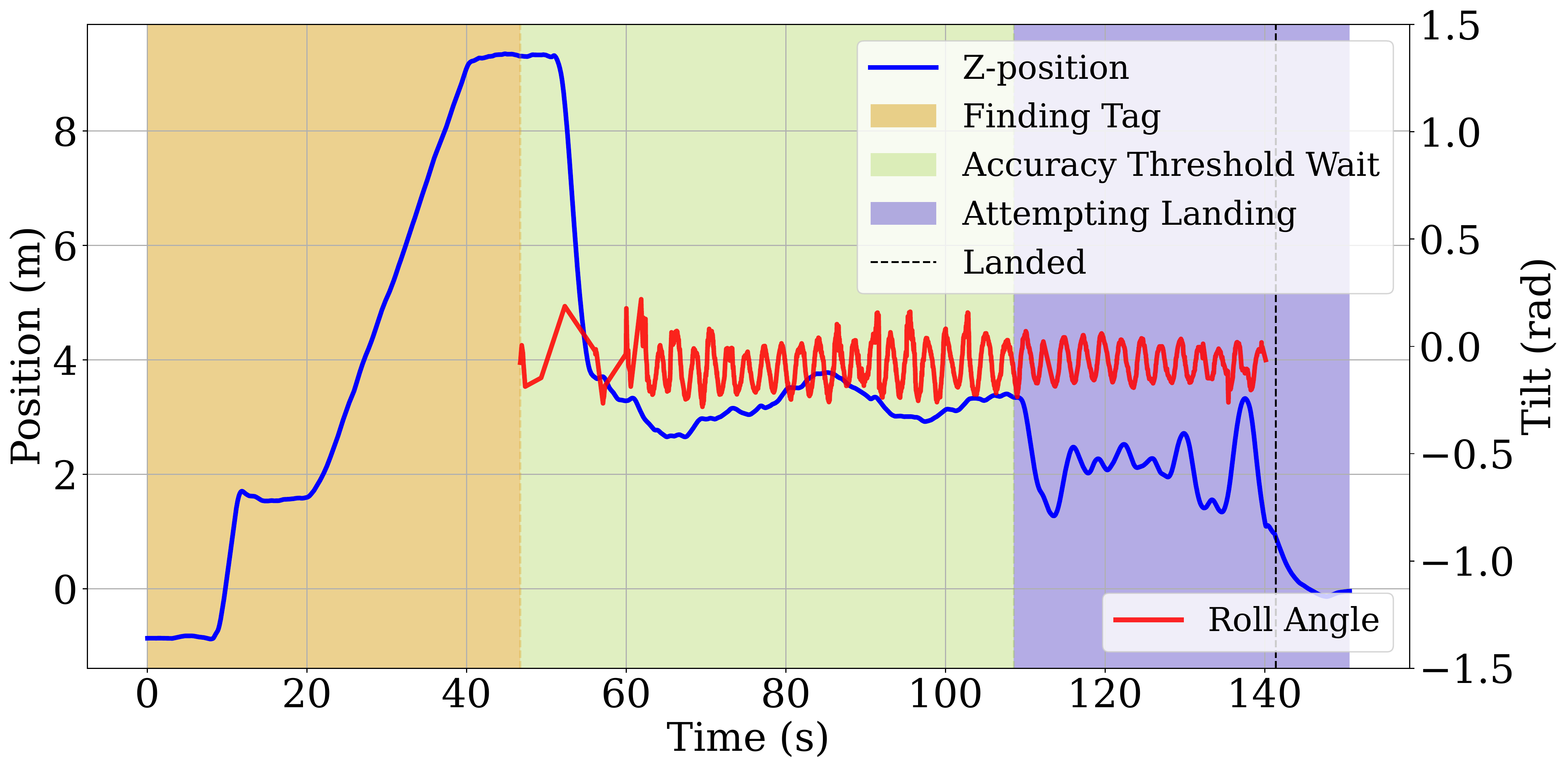}}

\caption{Plot of a selected real-world open-water experiment (\href{\medialink}{video}).}
\label{fig:real_experiment_landing}
\end{figure}

We demonstrate the real-world landing process through Figure \ref{fig:real_experiment_landing}. In these experiments, the \ac{uav} was able to land within $50$ seconds of acquiring the required \ac{fft} accuracy. This coincides with our findings in simulation experiments. Additionally, the tilt angles upon touchdown were less than $5^\circ$($0.09$~\text{rad}).

\section{Conclusion}

In this paper, we proposed an \ac{mpc} that enables a \ac{uav} to land autonomously on a tilting \ac{usv}. The \ac{mpc} employs a novel objective function and an online decomposition of the motion of the vessel in order to attempt and complete the landing during a near-zero tilt of the landing platform. We successfully demonstrated that we are able to model and predict the behaviour of the \ac{uav} and \ac{usv} without active communication between them. Further, we establish a novel approach for landing on the \ac{usv} using these predictions, which autonomously adjusts the relative altitude for the \ac{uav} to ensure that the landing occurs as close to the zero-tilt state of the landing deck as possible, increasing safeness of the landing phase and reducing impact forces on the landing \ac{uav}. In comparison to state-of-the-art approaches, we achieved significant improvement in the case of landing in demanding conditions with high waves and high winds without knowing the dimensions of the \ac{usv}.

\bibliographystyle{IEEEtran}
\bibliography{IEEEabrv,refs}

\end{document}